\immediate\write18{makeindex \jobname.nlo -s nomencl.ist -o \jobname.nls}

\documentclass[journal]{IEEEtran}

\usepackage{cite}
\usepackage{hyperref}

\ifCLASSINFOpdf
   \usepackage[pdftex]{graphicx}
   \graphicspath{{../pdf/}{../jpeg/}}
   \DeclareGraphicsExtensions{.pdf,.jpeg,.png}
\else
   \usepackage[dvips]{graphicx}
   \graphicspath{{../eps/}}
   \DeclareGraphicsExtensions{.eps}
\fi
\usepackage{amsmath}
\usepackage{array}

\usepackage{enumerate,amssymb,epsfig,euscript,indentfirst,authblk}
\usepackage{subcaption}
\usepackage{booktabs}
\usepackage{nccmath}
\usepackage{lipsum}
\usepackage{mathtools, cuted}
\usepackage{algpseudocode}
\usepackage{graphics}
\usepackage{algorithm}
\usepackage{nomencl}
\makenomenclature

\usepackage{multirow}
\usepackage{makecell}
\usepackage{etoolbox}

\usepackage{soul,color}
\usepackage{gensymb}

\begin{document}
%
\title{Automatic Operator-level Parallelism Planning for Distributed Deep Learning -- A Mixed-Integer Programming Approach}
%
%
%

\author{
        Ruifeng~She,
        Bowen~Pang,
        Kai~Li,
        Zehua~Liu,
        and~Tao~Zhong
\thanks{R. She, B. Pang, K. Li, Z. Liu, and T. Zhong are with Noah's Ark Lab, Huawei. E-mail: she.ruifeng@huawei.com.}
}

\maketitle

\begin{abstract}
As the artificial intelligence community advances into the era of large models with billions of parameters, distributed training and inference have become essential. While various parallelism strategies—data, model, sequence, and pipeline—have been successfully implemented for popular neural networks on main-stream hardware, optimizing the distributed deployment schedule requires extensive expertise and manual effort. 
Further more, while existing frameworks with most simple chain-like structures, they struggle with complex non-linear architectures. Mixture-of-experts and multi-modal models feature intricate MIMO and branch-rich topologies that require fine-grained operator-level parallelization beyond the capabilities of existing frameworks.

We propose formulating parallelism planning as a scheduling optimization problem using mixed-integer programming. We propose a bi-level solution framework balancing optimality with computational efficiency, automatically generating effective distributed plans that capture both the heterogeneous structure of modern neural networks and the underlying hardware constraints. In experiments comparing against expert-designed strategies like DeepSeek's DualPipe, our framework achieves comparable or superior performance, reducing computational bubbles by half under the same memory constraints.

The framework's versatility extends beyond throughput optimization to incorporate hardware utilization maximization, memory capacity constraints, and other considerations or potential strategies. Such capabilities position our solution as both a valuable research tool for exploring optimal parallelization strategies and a practical industrial solution for large-scale AI deployment.

\end{abstract}


\begin{IEEEkeywords}
Automated Scheduling, Distributed Computation, Mixed-Integer Programming. 
\end{IEEEkeywords}

%
\IEEEpeerreviewmaketitle

%
%
%
%

 

\section{Introduction}
\label{introduction}




\begin{figure}[t]
\centering
\begin{tabular}{c}
  \includegraphics[width=0.98\linewidth]{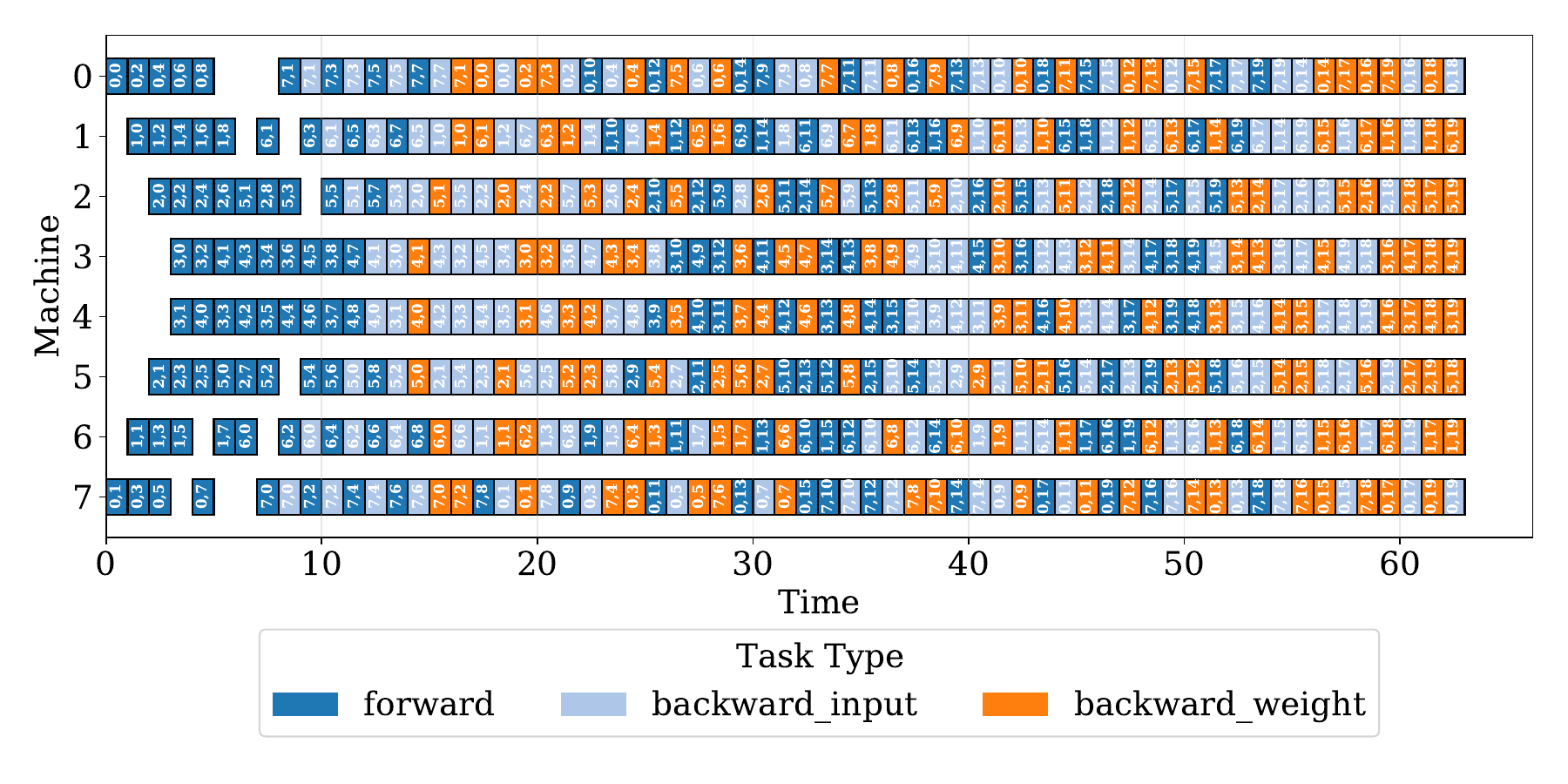} \\
  \small (a) Automatically searched schedule \\[0.5em]
  \includegraphics[width=0.98\linewidth]{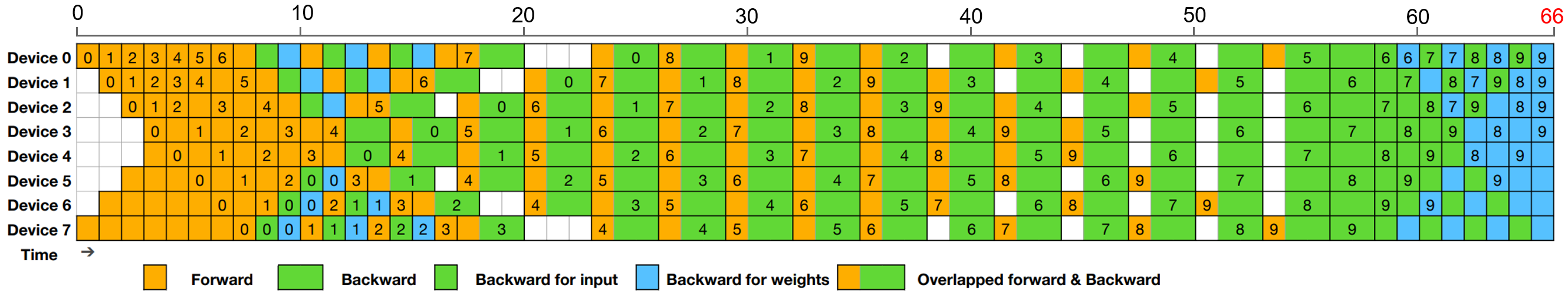} \\
  \small (b) DeepSeek v3's DualPipe (figure taken from \cite{deepseekai2024deepseekv3technicalreport})
\end{tabular}
\caption{Comparison of automatically searched parallelization schedule against expert-crafted schedule Dualpipe\cite{deepseekai2024deepseekv3technicalreport}. Colored blocks represent computation on each device, while white gaps indicate pipeline bubbles. The searched schedule achieved half of bubble counts as Dualpipe's, with similar scheduling pattern but different bubble distribution.}
\label{fig:optimal_vs_dualpipe}
\end{figure}

Large Language Models (LLMs) have emerged as a transformative force in artificial intelligence, revolutionizing capabilities across natural language processing tasks \cite{vaswani2017attention, brown2020language}. These models have demonstrated remarkable abilities in text generation, comprehension, and reasoning by scaling to unprecedented sizes, with state-of-the-art architectures now reaching trillions of parameters \cite{chowdhery2022palm, touvron2023llama}. The phenomenal emergence of LLMs has triggered exponential growth in their applications across industries, from conversational agents to content creation and sophisticated reasoning systems. However, this rapid expansion comes with substantial computational challenges as deploying larger models demands increasingly sophisticated hardware infrastructures and efficient computing strategies \cite{kaplan2020scaling}. Organizations seeking to leverage these powerful models face critical constraints in computational efficiency, as training and inference costs scale dramatically with model size. Single-device training and inference have become impractical for these massive models, making distributed computation not merely an optimization but a fundamental necessity \cite{rajbhandari2020zero, zheng2022alpa}.

\begin{figure}[h]
\centering
  \includegraphics[width=0.8\linewidth]{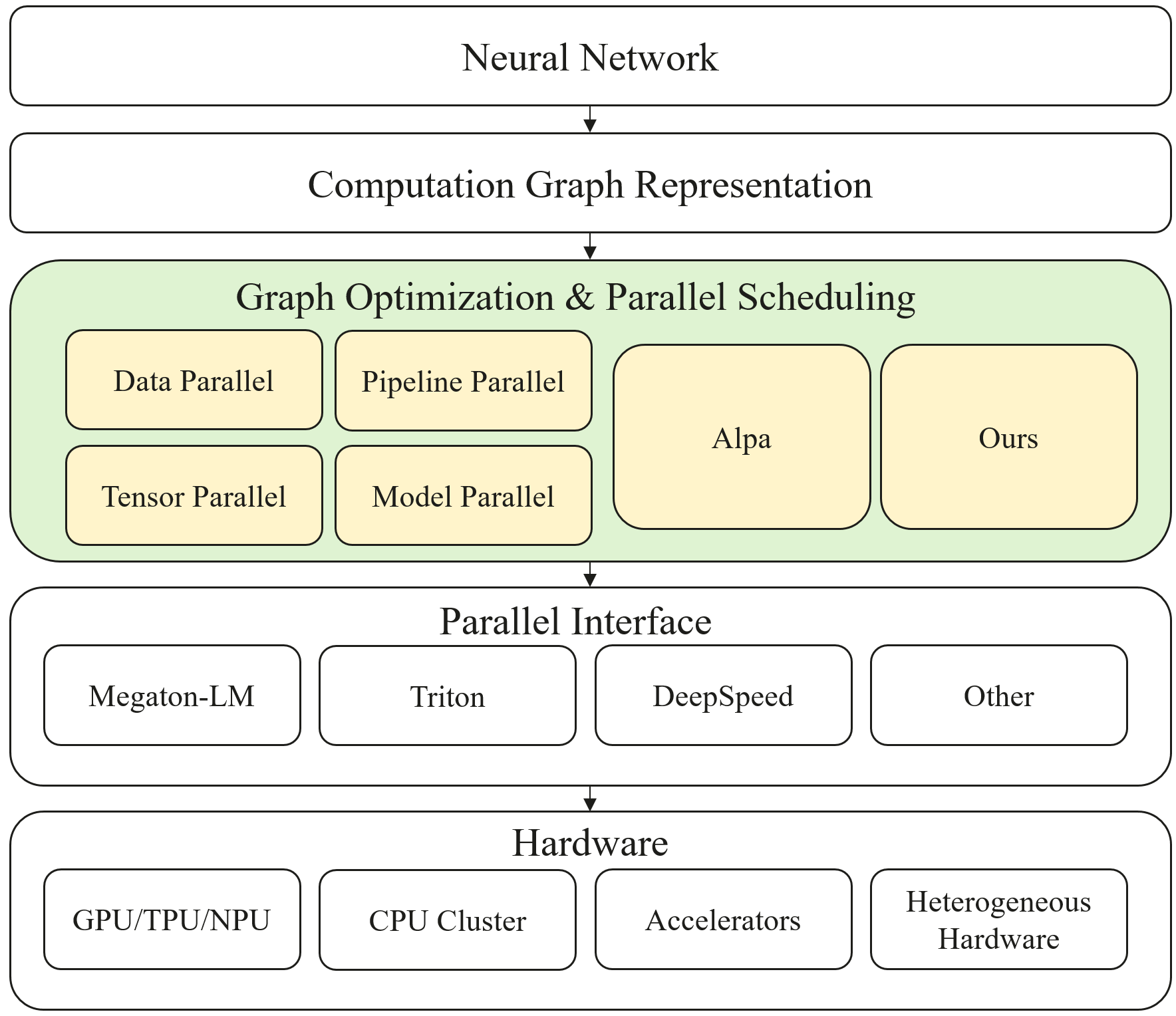}\\
  \caption{High-level illustration of parallel computation.}
  \label{fig:up_down_stream_position}
\end{figure}

To address these challenges, a typical top-to-bottom framework for employing parallel computation is illustrated in Figure~\ref{fig:up_down_stream_position}. The neural network module, encapsulating the architecture and learned parameters, can be represented as a graph, where nodes and edges capture the computational dependencies and data flow within the network. To facilitate efficient parallelization across multiple computing devices, a parallel schedule is designed to manage the communication and synchronization between nodes, ensuring efficient interaction and data exchange. The schedule is then recompiled through the optimized parallel interface into command sequences for the hardware level. Several pioneering companies have developed sophisticated systems to enable parallel training and deployment of neural networks. Microsoft's DeepSpeed offers a comprehensive suite of parallelism techniques including Zero Redundancy Optimizer (ZeRO), tensor, pipeline, and data parallelism that can be combined to enable training of extremely large models \cite{rasley2020deepspeed}. Similarly, Google's GShard and Nvidia's Megatron-LM represent other notable contributions that leverage model parallelism and sharding techniques to distribute computation efficiently across multiple devices \cite{lepikhin2020gshard, narayanan2021efficient}. While these approaches have pushed the boundaries of what's possible, they remain largely manual, requiring significant expertise to implement and tune for specific hardware configurations.

The current framework for designing parallel computation systems for LLMs involves significant manual effort and domain expertise. Engineering teams must navigate complex design spaces involving distributed computing architectures, memory constraints, communication patterns, and hardware-specific optimizations \cite{rajbhandari2020zero, shoeybi2019megatron}. A successful example is DeepSeek's DualPipe, featuring a hybrid bi-directional pipeline mechanism that optimizes both memory utilization and computational throughput \cite{deepseekai2024deepseekv3technicalreport}. Evidently, this process typically requires iterative experimentation and fine-tuning by specialists with deep understanding of both the underlying hardware and the specific characteristics of language models. The lack of an automatic designing method creates substantial barriers to entry for many organizations and researchers, limiting innovation and accessibility in the field \cite{narayanan2021efficient}. Furthermore, as models continue to grow in size and complexity, manual design approaches become increasingly impractical and error-prone. Recent frameworks orchestrate sophisticated rules and strategies to automate parallel deployment up to certain extents. However, these frameworks typically presume chain-like neural network structures and make specific simplifications, which may be limited in applicability or lead to suboptimal utilization for complex architectures.

In this paper, we introduce an automatic framework for designing parallelization strategies for LLM training and inference, fitting in the colored module in Figure~\ref{fig:up_down_stream_position}. We bridge the gap between theoretical scheduling optimization and practical distributed deep learning by formulating the parallelization problem as a mixed-integer program. By formalizing the parallelization strategy as a variant of job shop scheduling problem (JSSP), we enable automatic discovery of efficient execution plans tailored to specific hardware configurations and model architectures, considering various hardware constraints and performance objectives. We applied our proposed framework to optimize the training process of a neural network under memory constraints, and recovered parallelization plans achieving the same utility as expert-tailored strategy Dualpipe by Deepseek \cite{deepseekai2024deepseekv3technicalreport}. Furthermore, by continuing the search, we discovered plans achieving even better utility that halved the bubble counts than that of Dualpipe, as shown in Fig. \ref{fig:optimal_vs_dualpipe}, demonstrating the benefit of automating parallelization designing from an expanded solution space. Our framework represents a significant step toward automating the design of efficient distributed computation strategies for modern neural networks facilitating formal mathematical programming. By formalizing the parallelization problem within the well-established MIP methodology, we enable systematic exploration of the design space and identification of optimal or near-optimal solutions without relying exclusively on heuristics or expert intuition.

Our main contributions are summarized as follows:

\begin{itemize}
    \item We propose a comprehensive mixed-integer programming (MIP) formulation that captures the essential aspects of distributed neural network computation, including operator dependencies, device capabilities, and communication costs.
    \item We design a bi-level solution framework that balances optimality with computational efficiency, making the approach practical for real-world model sizes and hardware configurations.
    \item We conduct experimental validation of the effectiveness of our approach. Through an experiment to reproduce the Dualpipe scheduling strategy by DeepSeek, we demonstrate the capability of our framework to automatically discover comparable or potentially better performance than expert-designed strategies.
\end{itemize}

The remainder of the paper is structured as follows. A comprehensive literature review is given in section \ref{literature}. We present our problem formulation for automatically parallelization design as a MIP in section \ref{problem_formulation} along with a few extensions. We introduce our solution approach to solve the proposed model in section \ref{solution_method}. Numerical results are laid out in section \ref{numerical study} demonstrating how this framework can advance both research and practical deployment of large-scale deep learning models. Finally, we make conding remakrs in section \ref{conclusion}.

\section{Literature review}
\label{literature}

\subsection{Distributed Deep Learning Paradigms}

\begin{table*}[t]
    \centering
    \caption{Literature Review}
    \begin{tabular}{cccccccccc}
    \toprule
        Work & \makecell{Automatic \\ Scheduling} & \makecell{Theoretical \\ Bound} & \makecell{Makespan \\ Minimization} & \makecell{Throughput \\ Maximization} & \makecell{Memory \\ Management}  & \makecell{Nonlinear \\ Structure} & \makecell{Flexible \\ Parallelization} \\
    \midrule
        Megatron-LM\cite{shoeybi2019megatron} &  &  &  \checkmark &  &  &  &   \\
        DeepSpeed\cite{rasley2020deepspeed} &  &  & \checkmark  & \checkmark &  &  \checkmark &  \checkmark \\
        Alpa\cite{zheng2022alpa} & \checkmark &  & \checkmark  &  & \checkmark & \checkmark & \checkmark \\
        FlexFlow\cite{jia2018exploring} & \checkmark &  & \checkmark  & \checkmark &  & \checkmark & \checkmark \\
        Galvatron\cite{feng2023galvatron} & \checkmark &  & \checkmark  & \checkmark & \checkmark  &  & \checkmark \\
        GSPMD\cite{xu2021gspmd} & & & \checkmark & & \checkmark &   & \checkmark \\    
        PipeDream\cite{narayanan2021memory} & \checkmark & & \checkmark & \checkmark & \checkmark &  & \\   
        DualPipe\cite{deepseekai2024deepseekv3technicalreport} & & & \checkmark & \checkmark & \checkmark &  & \checkmark \\ 
    \midrule
        OURS & \checkmark & \checkmark & \checkmark & \checkmark & \checkmark & \checkmark & \checkmark \\
    \bottomrule
    \end{tabular}

    \label{table:literature_review}
\end{table*}

Researchers and practitioners have developed several parallelism strategies to address the challenges of distributed deep learning. These approaches can be broadly categorized into data parallelism, model parallelism, pipeline parallelism, and tensor parallelism, each with distinct characteristics and trade-offs \cite{li2020pytorch, shoeybi2019megatron}.

Data parallelism represents the most straightforward approach, where identical copies of the model operate on different data batches across devices, with periodic gradient synchronization \cite{li2020pytorch}. While conceptually simple and offering good scaling for computation-bound scenarios, data parallelism faces significant communication overhead as model sizes grow, limiting its effectiveness for the largest models \cite{rajbhandari2020zero}.

Model parallelism divides the neural network across devices, with each device responsible for specific layers or components \cite{shoeybi2019megatron}. This approach directly addresses memory constraints but introduces synchronization points that can lead to device underutilization. Tensor parallelism, a specialized form of model parallelism, splits individual operations (such as matrix multiplications) across devices, enabling more fine-grained distribution \cite{shoeybi2019megatron}.

Pipeline parallelism organizes computation into stages executed across different devices, with activations flowing between stages in a pipelined manner \cite{huang2019gpipe, narayanan2021memory}. This approach can improve device utilization but requires careful scheduling to minimize "bubbles" (idle periods) in the pipeline, particularly when dealing with complex model architectures \cite{fan2021dapple}.

Recent approaches have combined these strategies into hybrid parallelism schemes \cite{zheng2022alpa, narayanan2021efficient, li2021terapipe}, recognizing that no single strategy is optimal for all scenarios. These hybrid approaches typically rely on heuristics or expert knowledge to determine the parallelization strategy, often tailored to specific model architectures and hardware configurations.

The development of automated frameworks for designing and implementing distributed deep learning strategies has accelerated in recent years. We compare and tabulate a set of representative ones in Tabel~\ref{table:literature_review}. Systems like Megatron-LM \cite{shoeybi2019megatron} and DeepSpeed \cite{rasley2020deepspeed} have established foundational approaches for model and pipeline parallelism in transformer architectures. These frameworks often incorporate specific optimization techniques tailored to their target architectures but generally require manual configuration of parallelization strategies. More recent systems have aimed to automate aspects of this decision-making process. FlexFlow \cite{jia2018exploring} pioneered the use of search-based methods to explore parallelization strategies for neural networks, considering multiple dimensions of parallelism simultaneously. This approach was further expanded by GSPMD \cite{xu2021gspmd}, which uses a sharding propagation technique to automate tensor partitioning decisions. Alpa \cite{zheng2022alpa} introduced a hierarchical approach that separates intra-operator and inter-operator parallelism decisions, using a combination of dynamic programming and integer linear programming to search for efficient configurations. Unity \cite{unger2023unity} extended these ideas to heterogeneous clusters by incorporating hardware-aware cost modeling. Other systems like PipeDream \cite{narayanan2021memory} and Galvatron \cite{feng2023galvatron} have focused on specific aspects such as memory optimization, communication reduction, or heterogeneous resource allocation. 
Despite these advances, most existing frameworks either rely on simplified models of the underlying computation graphs, focus on specific model architectures, or employ heuristic search methods that may yield suboptimal solutions for complex, non-linear neural network topologies. Additionally, many approaches encounter scalability challenges when applied to the largest models, highlighting the need for more principled optimization frameworks that can efficiently navigate the vast design space of distributed deep learning configurations.

\subsection{Challenges in Modern Model Architectures}
While existing parallelism approaches have proven effective for models with relatively simple, sequential structures (such as standard transformers), they face significant challenges when applied to modern architectures with more complex topologies \cite{lepikhin2020gshard, fedus2022switch, radford2021learning}. Mixture-of-expert models \cite{fedus2022switch, lepikhin2020gshard}, for example, introduce conditional branching patterns that create imbalanced computation paths. Multi-modal architectures \cite{radford2021learning} combine different network structures with varying computational characteristics. These models feature Multiple-Input Multiple-Output (MIMO) patterns and branch-rich topologies that fundamentally complicate the parallelization process.

The standard approaches to pipeline parallelism, which typically partition models into sequential stages, struggle with these non-linear structures. Efficiently distributing computation for such models requires operator-level decisions that consider the unique characteristics of each component and their interactions \cite{narayanan2021efficient, zheng2022alpa}. Manual design of parallelization strategies for these complex architectures demands significant expertise and experimentation, often leading to suboptimal solutions due to the vast design space \cite{jia2019beyond}. Furthermore, hardware heterogeneity introduces additional complexity. Modern AI accelerators vary in memory capacity, compute capability, and interconnect bandwidth. Effective parallelization must account for these differences to achieve optimal performance, further complicating the design process.

\subsection{Scheduling Problems in Distributed Computing} 
The challenge of optimally distributing neural network computation across multiple devices bears strong resemblance to classical scheduling problems, particularly the job shop scheduling problem (JSSP) \cite{pinedo2012scheduling}. In JSSP, a set of jobs must be processed on a set of machines, with each job comprising a sequence of operations with specified processing times. The objective is typically to minimize the total completion time (often referred to as the makespan) while respecting precedence constraints and machine availability.
Such a resemblance is particularly obvious when jointly considering model and pipeline parallelism for neural networks. The forward and backward passes through network layers can be viewed as operations with specific processing requirements and precedence constraints, while accelerator devices represent the machines on which these operations must be scheduled \cite{narayanan2021memory, fan2021dapple}. The goal of minimizing training or inference time corresponds directly to the makespan minimization objective in JSSP.
On the other hand, data parallelism poses a top-level load-balancing assignment problem, aiming to partition the machines such that the job loads are equally assigned to each machine and are completed simultaneously.
Tensor parallelism instead can be viewed as a bottom-level sub-problem of JSSP, where a single operation is processed across multiple machines.

Several researchers have recognized this connection and applied scheduling techniques to distributed deep learning. \cite{fan2021dapple} formulated pipeline parallelism as a scheduling problem and developed heuristic solutions. \cite{narayanan2021memory} proposed PipeDream, which uses dynamic programming to find efficient pipeline configurations. \cite{zheng2022alpa} developed a two-level approach that combines intra-operator and inter-operator parallelism through hierarchical search. \cite{pang2025hybrid} formulate the inference serving system as an online scheduling problem and provided heuristics based on decomposition method. However, these approaches typically employ heuristics or simplified models due to the computational complexity of the underlying scheduling problem, which is known to be NP-hard \cite{pinedo2012scheduling}. They also often focus on specific model architectures or parallelism strategies rather than providing a general framework applicable to diverse scenarios. Mixed-integer programming (MIP), on the other hand, offers a powerful framework for modeling and solving scheduling problems with complex constraints \cite{pochet2006production}. While MIP approaches have been successfully applied in various manufacturing and service scheduling contexts \cite{manne1960job, ku2016mixed}, their application to distributed deep learning has been limited, primarily due to scalability concerns for large models.

\section{Problem formulation}
\label{problem_formulation}

Consider a neural network consisting of a set of operations, each requiring certain time to execute, that all of which must be executed to complete the inference or training task. Some operations produce intermediate outputs that are required as inputs by downstream operations, hence must be executed in order, while operations do not have such an dependency can be executed in any order. The operations are executed on cluster of computation resources, consisting of a set of machines (e.g., GPUs or NPUs) interconnected via communication channels, through which the data flow between operations can be transferred. Meanwhile, executing each operation requires allocating some space on the machine (i.e., memory) to store the output (e.g., activations) and the asset used by the operation (e.g.,  the weights and bias of a hidden layer). In addition, we assume the following assumptions:
\begin{itemize}
    \item All machines and communication channels are available at the beginning.
    \item Each machine can process at most one operation at any time, and each communication channel can process one communication task at any time\footnote{Here we assume single-thread machines for simplicity. One can model multi-thread machines by grouping multiple machines and form a sub-cluster accordingly.}.
    \item Operations and communication tasks cannot be interrupted once started (i.e., non-preemptive).
\end{itemize}
Such a problem falls into the class of Flexible Distributed Job Shop Scheduling Problems (FDJSSP), where \textit{flexible} indicates that the order of executing some operations can be interchanged, and \textit{distributed} indicates that each operation can be executed on one of many candidate machines.

Following the convenient representation of the disjunctive graph model in the literature of JSSP, we use a directed acyclic graph (DAG) $G_m(\mathcal{I}, \mathcal{B})$ to represent the computation graph of a neural network, where $\mathcal{I}$ denotes the set of operations, and $\mathcal{B}$ denotes the data dependency between operations. Similarly, we use a directed graph $G_h(\mathcal{J}, \mathcal{C})$ to denote the computation cluster, where $\mathcal{J}$ denotes the devices and $\mathcal{C}$ denotes the communication channels. Note that $G_h$ is likely cyclic due to the presence of bi-directional communication channels in most modern hardware. Throughout this paper, we will use upper-case symbols to denote known constant parameters, and lower-case symbols for decision variables and indices. We denote $D^{\text{o}}_i$ as the execution duration required for completing operation $i$, and $D^{\text{c}}_{i_1,i_2}$ as that for the communication task from operation $i_1$ to $i_2$ \footnote{Here we assume that the machines and communication channels are homogeneous for simplicity. The proposed model can be easily extended to heterogeneous situations by including the machine dimension relating to the operation allocation decisions.}. For simplicity, we assume the most memory-consuming asset to store are the weights of the operations and neglect the rest, and denote $A_i$ and $W_i$ as the memory requirement to store the output and the weight of operation $i$. Note that $A_i$ can be either positive, negative or 0, indicating an increasing, decreasing or unchanged memory level after executing the operation. For the communication channel, flow control and congestion management are adopted in most modern multi-thread devices, hence we assume that any arbitrarily large dataflow can be sent/received over longer communication duration without causing memory outflow. The automatic design of parallelization strategy thus aims to construct a valid arrangement (i.e., not exceeding the memory limit of any machine) for the operations onto machines that optimizes certain metric, such as the total completion time (often referred to as the makespan) of the neural network, or the throughput of a batch of requests, or many others.

\begin{table*}
    \centering
    \caption{Notation Table: Sets, Parameters and Decision Variables}
    \begin{tabular}{cl}
    \toprule
        Sets    & Description \\
    \midrule
        $\mathcal{I}$  & Set of operations. \\
        $\mathcal{B}$  & Set of data-dependent operations. \\
        $\mathcal{J}$  & Set of machines. \\
        $\mathcal{C}$  & Set of communication channels. \\
    \midrule
        Parameters          & \\
    \midrule
        $D^{\text{o}}_i, \forall i \in \mathcal{I}$             & Duration of operation $i$. \\
        $D^{\text{c}}_{i_1,i_2}, \forall (i_1,i_2) \in \mathcal{B}$             & Duration of communication task $(i_1, i_2)$. \\
        $W_i, \forall i \in \mathcal{I}$             & Memory requirement of operation $i$. \\
        $A_i, \forall i \in \mathcal{I}$             & Memory adjustment after operation $i$. \\
    \midrule
        Decision Variables      & \\
    \midrule
        $x_{i,j} \in \{0,1\}, \forall i \in \mathcal{I}, j \in \mathcal{J}$                      & Allocation of operation $i$ to machine $j$.  \\
        $y_{i_1,i_2} \in \{0,1\}, \forall (i_1, i_2) \in \mathcal{I} \times \mathcal{I}$         & Precedence of operation $i_1$ before $i_2$. \\
        $s_i, \forall i \in \mathcal{I}$      & Start time of operation $i$. \\
        $e_i, \forall i \in \mathcal{I}$      & End time of operation $i$. \\
        $t_{i_1,i_2}, \forall (i_1, i_2) \in \mathcal{I} \times \mathcal{I}$        & Slack time between operations $i_1$ and $i_2$. \\
        $z_{i_1,i_2,j_1,j_2} \in \{0,1\}, \forall (i_1, i_2) \in \mathcal{B}, (j_1,j_2) \in \mathcal{J} \times \mathcal{J}$        & Allocation of communication task $(i_1, i_2)$ to channel $(j_1,j_2)$. \\
        $w_{i_1,i_2,i_3,i_4} \in \{0,1\}, \forall (i_1, i_2), (i_3, i_4) \in \mathcal{B} \times \mathcal{B}$        & Precedence of communication task $(i_1, i_2)$ before $(i_3,i_4)$. \\
        $c_{i_1,i_2}, \forall (i_1, i_2) \in \mathcal{B}$       & Start time of communication task $(i_1, i_2)$. \\
        $d_{i_1,i_2}, \forall (i_1, i_2) \in \mathcal{B}$       & End time of communication task $(i_1, i_2)$. \\
        $u_{i_1,i_2} \in \{0,1\}, \forall (i_1, i_2) \in \mathcal{I} \times \mathcal{I}$     & Immediate precedence of operation $i_1$ before $i_2$. \\
        $m^-_i, \forall i \in \mathcal{I}$     & Memory level before operation $i$. \\
        $m^+_i, \forall i \in \mathcal{I}$     & Memory level after operation $i$. \\
    \bottomrule
    \end{tabular}
    \label{notation table}
\end{table*}

The FDJSSP minimizing the makespan can be formulated as the following MIP:

\begin{align}
    \label{obj:z}
    & \min z \\
    & \noindent \nonumber \text{s.t.} \\
    \label{cst:makespan}
    & z \geq e_i, \forall i \in \mathcal{I}, \\
    \label{cst:duration_rel}
    & e_i - s_i = D^{\text{o}}_i, \forall i \in \mathcal{I}, \\
    \label{cst:slack_def}
    & s_{i_1} - s_{i_2} - t_{i_1,i_2} = 0, \forall i_1 \neq i_2 \in \mathcal{I}, \\
    \label{cst:bom_def}
    & t_{i_1,i_2} \geq 0, \forall (i_1,i_2) \in \mathcal{B}, \\
    \label{cst:single_assignment}
    & \sum_{j \in \mathcal{J}} x_{i,j} = 1, \forall i \in \mathcal{I}, \\
    \label{cst:non_overlap}
    \begin{split}
        & -M (3 - y_{i_1,i_2} - x_{i_1j} - x_{i_2j}) \\ & + e_{i_1} - s_{i_2} \leq 0, \forall i_1 \neq i_2 \in \mathcal{I}, j \in \mathcal{J}, 
    \end{split} \\
    \label{cst:v_def}
    & y_{i_1,i_2} + y_{i_2i_1} = 1, \forall i_1 \neq i_2 \in \mathcal{I}, \\
    \label{cst:comm_loc_def}
    \begin{split}
    & z_{i_1,i_2,j_1,j_2} = x_{i_1j_1} \cdot x_{i_2j_2}, \\ & \forall (i_1,i_2) \in \mathcal{B}, (j_1,j_2) \in \mathcal{C}    
    \end{split} \\
    \label{cst:comm_channel_def}
    & z_{i_1,i_2,j_1,j_2} \leq 0, \forall (j_1,j_2) \notin \mathcal{C}, \\
    \label{cst:comm_duration_rel}
    & d_{i_1,i_2} - c_{i_1,i_2} \geq D^{\text{c}}_{i_1,i_2}, \forall (i_1,i_2) \in \mathcal{B}, \\
    \label{cst:comm_ed_rel}
    & s_{i_2} - d_{i_1,i_2} \geq 0, \forall (i_1,i_2) \in \mathcal{B}, \\
    \label{cst:comm_st_rel}
    & e_{i_1} - c_{i_1,i_2} \leq 0, \forall (i_1,i_2) \in \mathcal{B}, \\
    \label{cst:comm_non_overlap}
    \begin{split}
        & M (3 - w_{i_1,i_2,i_3,i_4} - z_{i_1,i_2,j_1,j_2}- z_{i_3i_4j_1j_2}) + c_{i_3i_4} \\ 
        &  - d_{i_1,i_2} \geq 0,\forall (i_1, i_2) \neq (i_3, i_4) \in \mathcal{B}, (j_1, j_2) \in \mathcal{C},
    \end{split} \\
    \label{cst:v_c_def}
    & w_{i_1,i_2,i_3,i_4} + w_{i_3i_4i_1i_2} = 1,  \forall (i_1, i_2) \neq (i_3, i_4) \in \mathcal{B},\\
    \label{cst:mem_init}
    & M(1 - x_{i,j}) + m^-_i \geq \sum_{i} W_i x_{i,j}, \forall i \in \mathcal{I}, \forall j \in \mathcal{J},\\
    \label{cst:mem_rel_1}
    & M(1 - u_{i_1,i_2}) + m^-_{i_2} - m^+_{i_1} \geq 0,  \forall i_1 \neq i_2, i_1, i_2 \in \mathcal{I},     \\
    \label{cst:mem_rel_2}
    & -M(1 - u_{i_1,i_2}) + m^-_{i_2} - m^+_{i_1} \leq 0,  \forall i_1 \neq i_2 \in \mathcal{I},\\
    \label{cst:mem_def}
    & m^+_i - m^-_i - A_i = 0, \forall i \in \mathcal{I},
\end{align}

where $M$ denotes a large positive number. In the above, Eqs. \ref{obj:z}-\ref{cst:makespan} state that the objective is to minimize the makespan, which equals the ending time of the last operation; Eqs. \ref{cst:duration_rel} relate the starting time, ending time, and the execution duration of an operation; Eqs. \ref{cst:slack_def} define the slack time between operations, and Eqs. \ref{cst:bom_def} enforces the ordering between data-dependent operations; Eqs. \ref{cst:single_assignment} guarantee that each operation is allocated to exactly one machine; Eqs. \ref{cst:non_overlap}-\ref{cst:v_def} together state that if any two operations are executed on the same machine, their execution time must not overlap; Eqs. \ref{cst:comm_loc_def} relates the operation-machine allocation to the corresponding communication-channel allocation\footnote{Eqs. \ref{cst:comm_loc_def} is written as quadratic constraints for conciseness, and they can be solved by modern solvers such as Gurobi. Such constraints can be linearized using techniques suggested in \cite{sherali2010reformulation}}; Eqs. \ref{cst:comm_channel_def} negates communication between machines that are not connected by a communication channel; Eqs. \ref{cst:comm_duration_rel} relate the starting time, ending time, and the execution duration of a communication task; Eqs. \ref{cst:comm_ed_rel}-\ref{cst:comm_st_rel} relate the starting/ending of a communication task with its up/downstream dependent task, respectively; Eqs. \ref{cst:comm_non_overlap}-\ref{cst:v_c_def} together state that if any two communication tasks are executed on the same communication channel, their execution time must not overlap; Eqs. \ref{cst:mem_init} initializes the memory level for operations on each machine; Eqs. \ref{cst:mem_rel_1}-\ref{cst:mem_def} simulate the change of memory level on the machine.

\section{Solution Method}
\label{solution_method}


The proposed MIP formulation provides a comprehensive and general framework for addressing the automatic parallelization problem. However, due to the combinatorial nature of distributed scheduling problems, the computational complexity grows significantly as the number of operations increases, making it challenging to obtain solutions within a limited time budget. To address this challenge, we first introduce a heuristic approach that pre-merges certain operations, thereby reducing the problem size and achieving a balance between solution optimality and computational cost.

Inspired by priority dispatch rule-based algorithms, our heuristic iteratively identifies pairs of nodes for merging based on a set of predefined rules. The core of the algorithm consists of two key functions: $GetCandidateEdge(\mathcal{I}, \mathcal{B})$ and $GetCandidateNonEdge(\mathcal{I}, \mathcal{B})$. These functions suggest candidate node pairs for merging, with the distinction that $GetCandidateEdge$ focuses on pairs connected by an edge in the graph, while $GetCandidateNonEdge$ considers pairs without a direct edge. During each function call, topologically redundant edges, defined as $\{(i_1,i_2) \in \mathcal{B} \mid \exists i_3:(i_1,i_3) \in \mathcal{B}, (i_3,i_2) \in \mathcal{B}\}$, are temporarily removed. This step is crucial to reveal structurally important edges and prevent the formation of cycles after node merging.

The function $GetCandidateEdge$ identifies edges where the head node has a single input and the tail node has a single output, subject to predefined thresholds on operation duration and memory requirements. Such node pairs are deemed "safe" for merging, as their combination does not increase the critical path length of the graph. Conversely, merging nodes that are not directly connected is considered "unsafe," as it may inadvertently extend the critical path length. To mitigate this risk, $GetCandidateNonEdge$ imposes stricter thresholds to avoid creating overly large operations that could become performance bottlenecks.

The function $MergeNodes$ is responsible for merging the selected node pairs and updating the graph structure accordingly. It ensures that the new node is correctly connected to the rest of the graph based on the original connectivity, while also updating node and edge attributes. For instance, the duration and memory requirements of the newly merged node are computed as the sum of the corresponding attributes of the constituent nodes. The pseudo-code of the algorithm is provided in Algorithm \ref{algorithm1}.

We acknowledge that this heuristic is inherently greedy. Effectively simplifying the graph $G_m$ requires solving a graph partitioning problem with potentially complex constraints, which we leave as a direction for future research. Once $G_m$ is reduced to a manageable size—typically fewer than 100 nodes—we employ a commercial MIP solver to solve the proposed model efficiently.


\begin{algorithm}[h]
\caption{Heuristic to merge operations in JSSP}
\label{algorithm1}
\begin{algorithmic}
\Require $G_m(\mathcal{I}, \mathcal{B}), N$ 

\While{True}
    \While{{GetCandidateEdge}$(\mathcal{I}, \mathcal{B})$ returns} \Comment{Loop over to merge edges}
        \State $(node_1, node_2) \gets $ {GetCandidateEdge}$(\mathcal{I}, \mathcal{B})$
        \State $\mathcal{I}, \mathcal{B} \gets $ {MergeNodes}$(\mathcal{I}, \mathcal{B}, node_1, node_2)$
        \State Break if $|\mathcal{I}|\leq N$
    \EndWhile
    \While{{GetCandidateEdge}$(\mathcal{I}, \mathcal{B})$ returns} \Comment{Loop over to merge non-edges}
        \State $(node_1, node_2) \gets$ GetCandidateNonEdge$(\mathcal{I}, \mathcal{B})$
        \State $\mathcal{I}, \mathcal{B} \gets $ {MergeNodes}$(\mathcal{I}, \mathcal{B}, node_1, node_2)$
        \State Break if $|\mathcal{I}|\leq N$
    \EndWhile
\EndWhile
\end{algorithmic}
\end{algorithm}






\section{Numerical Study}
\label{numerical study}

In this section, we present the numerical results obtained from our proposed framework. We employed the commercial solver Gurobi to solve the MIP. The framework is implemented in Python on PC with an Intel Core i7 processor, 32 GB of RAM. 

\subsection{Reproducing and Enhancing Dualpipe Strategy}

We first validate our framework's ability to rediscover expert-designed parallelization strategies and potentially improve upon them. We compare our approach with Dualpipe, a pipeline parallelism technique implemented in DeepSeek V3. Dualpipe represents a state-of-the-art manual parallelization strategy designed to balance memory constraints and computational efficiency for large language models.

Our first objective was to verify if our mathematical programming framework could automatically derive the Dualpipe strategy when given identical constraints. We follow the technical report of DeepSeek V3 and assume a neural network divided into stages, where the number of stages = the number of machines = the pipeline parallelization rank, denoted by $PP$. For simplicity, we assume that the execution time of forward, backward for input, and backward for weights of each stage all equals to one unit of time, denoted by $T_f=T_i=T_w=1$, and that of a full backward chunk equals $T_b = T_i + T_w = 2$. We formulated the optimization problem using Dualpipe's memory bound, which allocates memory on each device equal to $2 \times$ model parameters plus ($PP$ + 1) activation memory. This specific memory allocation supports Dualpipe's characteristic forward-backward scheduling while maintaining activation checkpoints. Additionally, we set a known primal bound on overall makespan of the sum of execution time of all stages plus the bubble size, which equals to $(PP / 2 - 1) \times (T_f + 2 T_b- 3 T_w) = 2(PP/2-1)$, forcing the solving process of the MIP to terminate when reaching the same performance.

\begin{figure}[t]
\centering
\begin{tabular}{c}
  \includegraphics[width=0.98\linewidth]{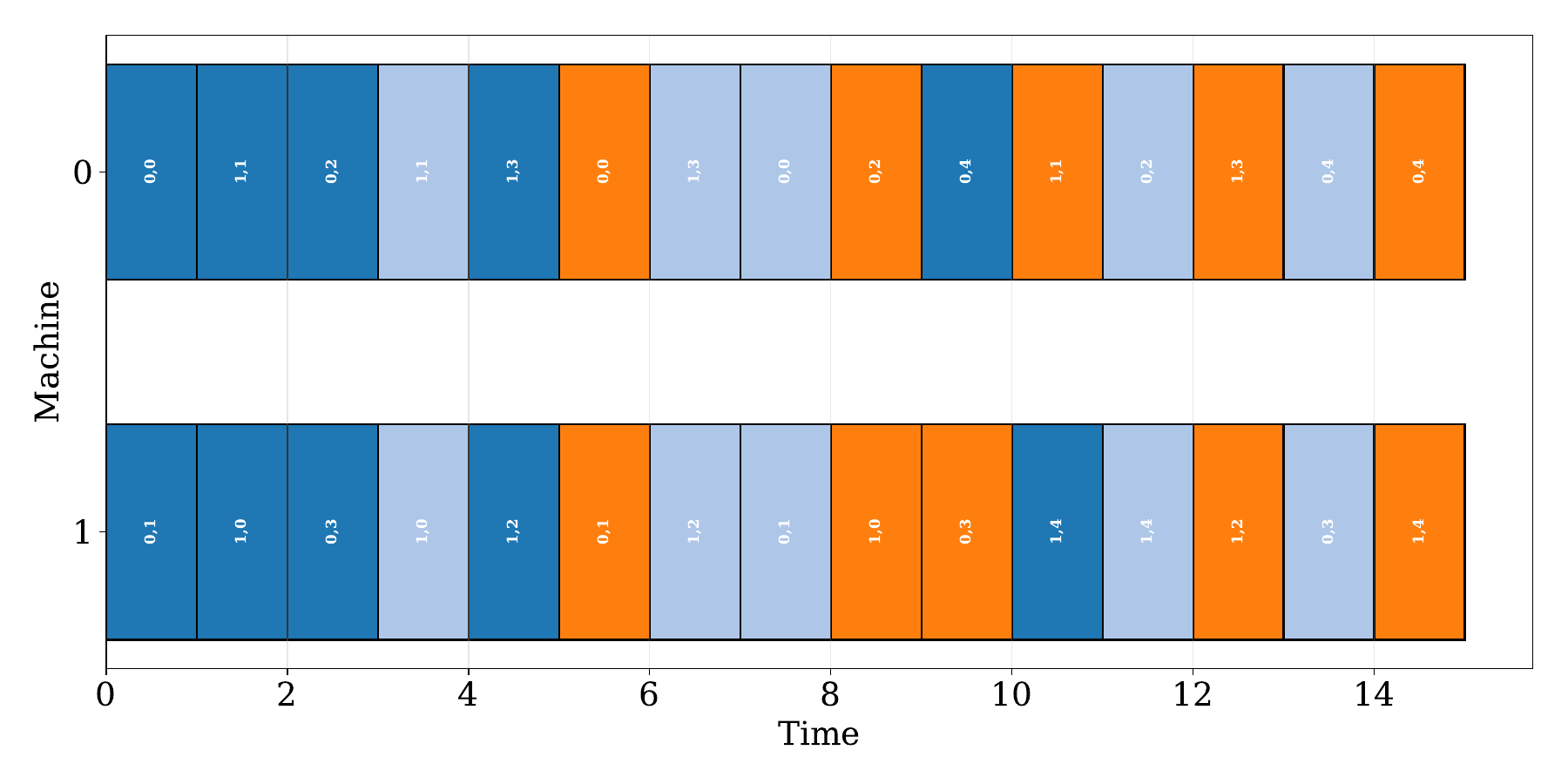} \\
  \small (a) PP-rank = 2 \\[0.5em]
  \includegraphics[width=0.98\linewidth]{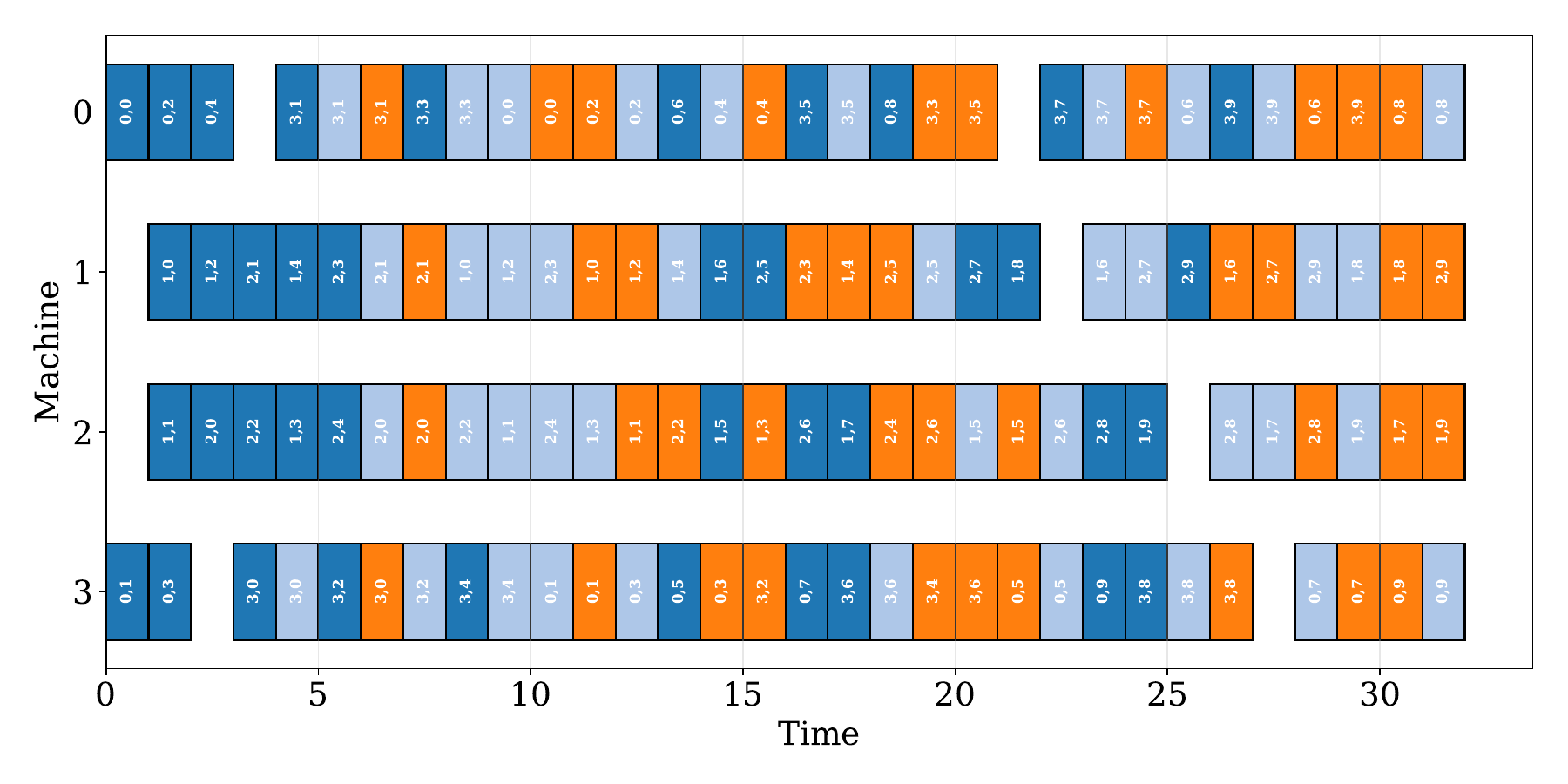} \\
  \small (b) PP-rank = 4 \\[0.5em]
  \includegraphics[width=0.98\linewidth]{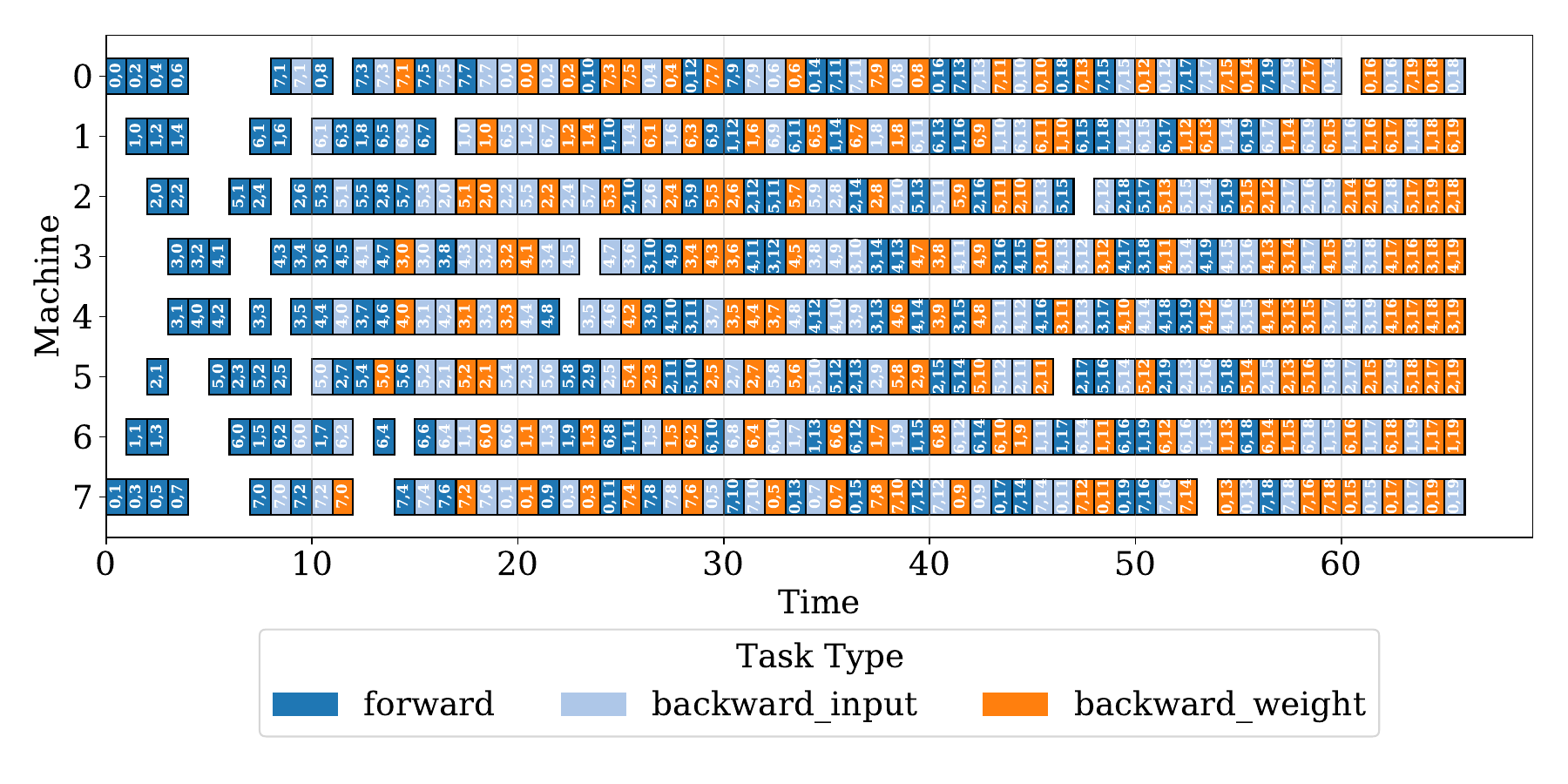} \\
  \small (c) PP-rank = 8
\end{tabular}
\caption{Gantt charts of parallelization strategies automatically generated by our framework under identical memory constraints as Dualpipe. Colored blocks represent computation on each device, while white gaps indicate pipeline bubbles. The solutions achieved a bubble ratio to Dualpipe's scheduling pattern, with similar pipeline bubble distribution.}
\label{fig:dualpipe-reproduction}
\end{figure}

Figure~\ref{fig:dualpipe-reproduction} presents Gantt charts of our optimizer's solutions across three different pipeline parallelism configurations ($PP$ = 2, 4, and 8). The horizontal axis represents time, while each row shows operations scheduled on a single device. Colored blocks indicate computation, and the gaps between operations represent pipeline bubbles—idle periods where devices wait due to dependencies. Evidently, our framework successfully reproduces Dualpipe's parallelization strategy across all configurations. The computational boundaries between pipeline stages and the resulting bubble patterns closely match those of the manually designed Dualpipe approach. The exact scheduling of forward and backward passes slightly differ from that of Dualpipe, which is expected because both solutions are equivalent from the perspective of the MIP. Interestingly, the schedules manifest a bi-directional interweaving pattern similar to Dualpipe without explicit arrangement. This result demonstrates our framework's ability to automatically discover sophisticated parallelization strategies through mathematical optimization, potentially eliminating weeks of expert engineering effort.

\begin{figure}[t]
\centering
\begin{tabular}{c}
  \includegraphics[width=0.98\linewidth]{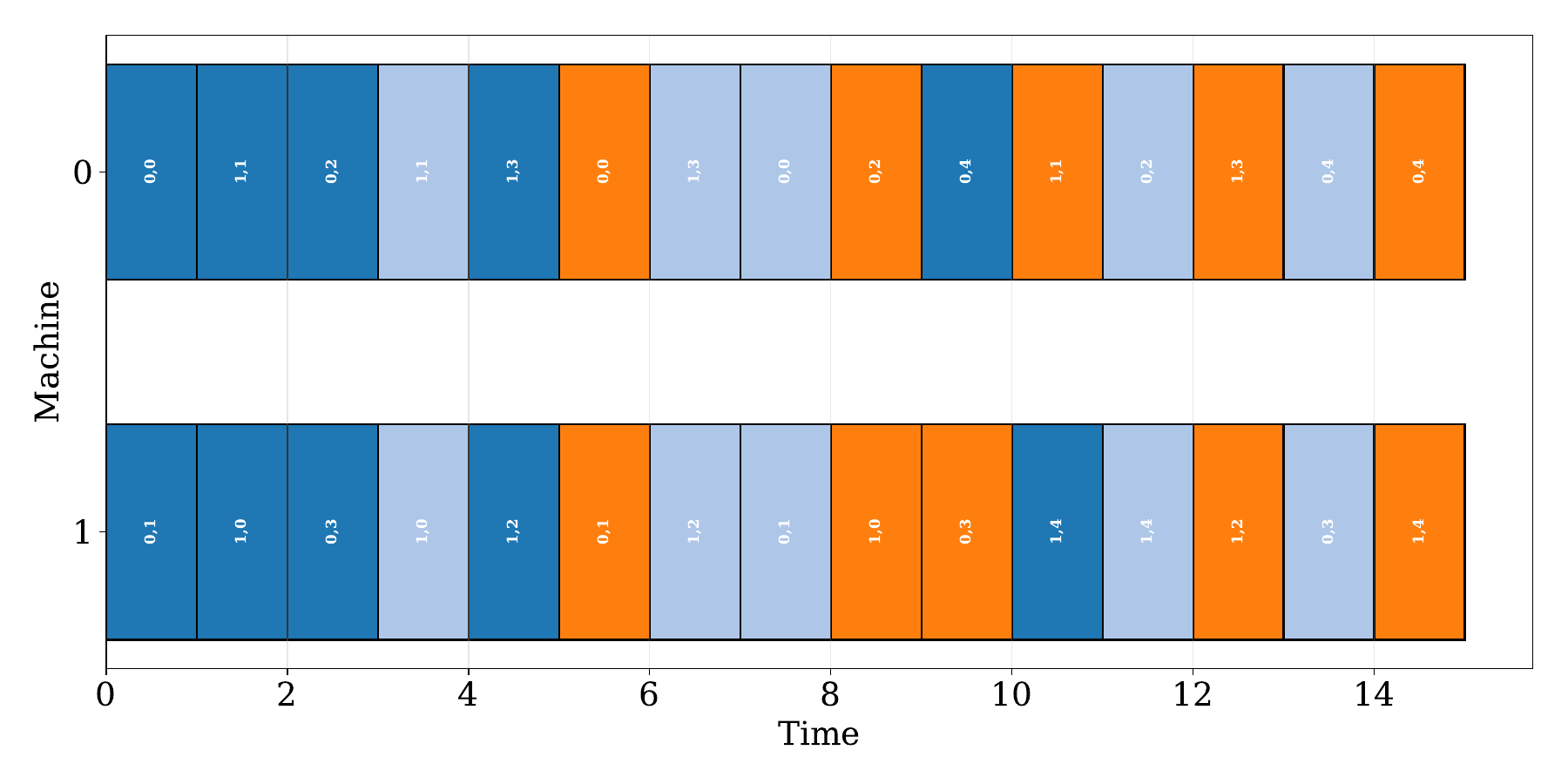} \\
  \small (a) PP-rank = 2, relaxed memory \\[0.5em]
  \includegraphics[width=0.98\linewidth]{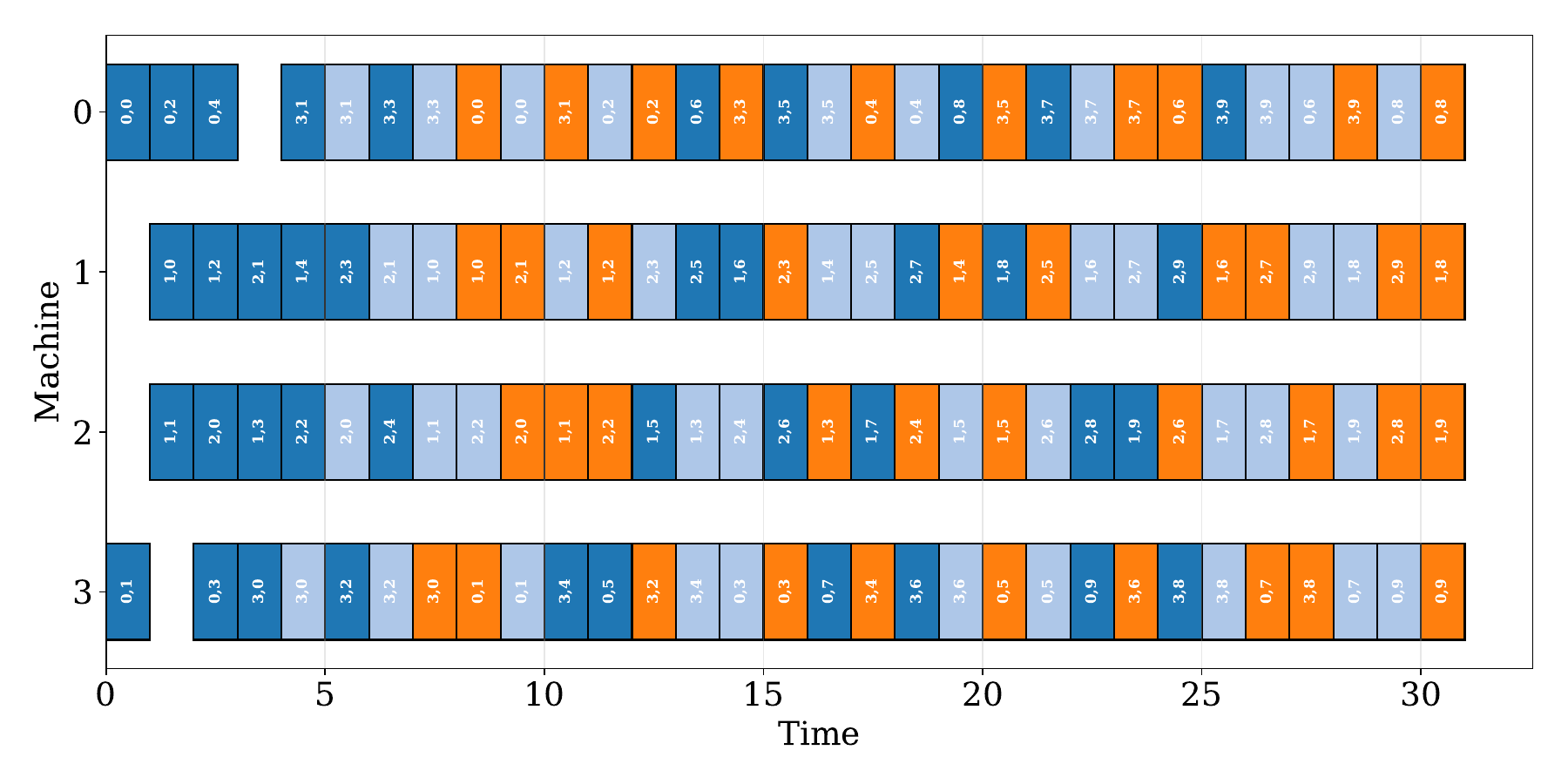} \\
  \small (b) PP-rank = 4, relaxed memory \\[0.5em]
  \includegraphics[width=0.98\linewidth]{figs_0312//dualpipe_reproduce_free_bound_8.pdf} \\
  \small (c) PP-rank = 8, relaxed memory
\end{tabular}
\caption{Gantt charts of enhanced parallelization strategies by continuing searching after reaching the utility by Dualpipe. The optimization framework discovers more efficient scheduling patterns, reducing pipeline bubbles by approximately 50\% compared to the strict Dualpipe memory configuration.}
\label{fig:dualpipe-better}
\end{figure}

Then, we relax the constraint on the primal bound and let the MIP solver further search the solution space. Figure~\ref{fig:dualpipe-better} presents Gantt charts of the scheduling at convergence, reaching a lower bubble size of $(PP/2-1)$ across all pipeline depths, which is half of that achieved by Dualpipe, leading to substantially improved hardware utilization and reduced end-to-end execution time.. It is worth noting that the new solution is found under the same memory bound constraints and without any manual adjustment, demonstrating the capability of the framework to automatically discover potentially better scheduling.

\begin{figure}[t]
\centering
\begin{tabular}{c}
  \includegraphics[width=0.98\linewidth]{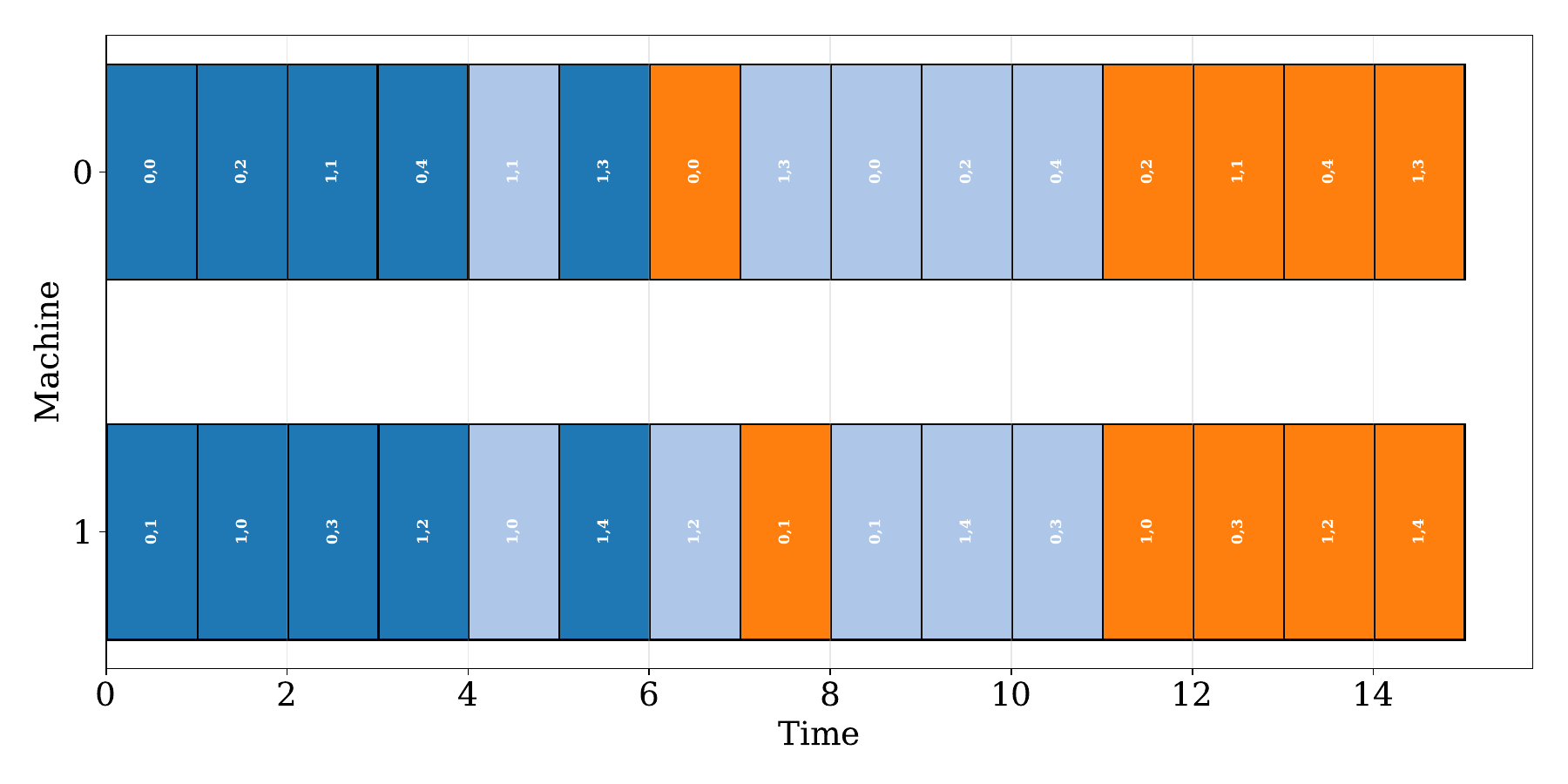} \\
  \small (a) PP-rank = 2 \\[0.5em]
  \includegraphics[width=0.98\linewidth]{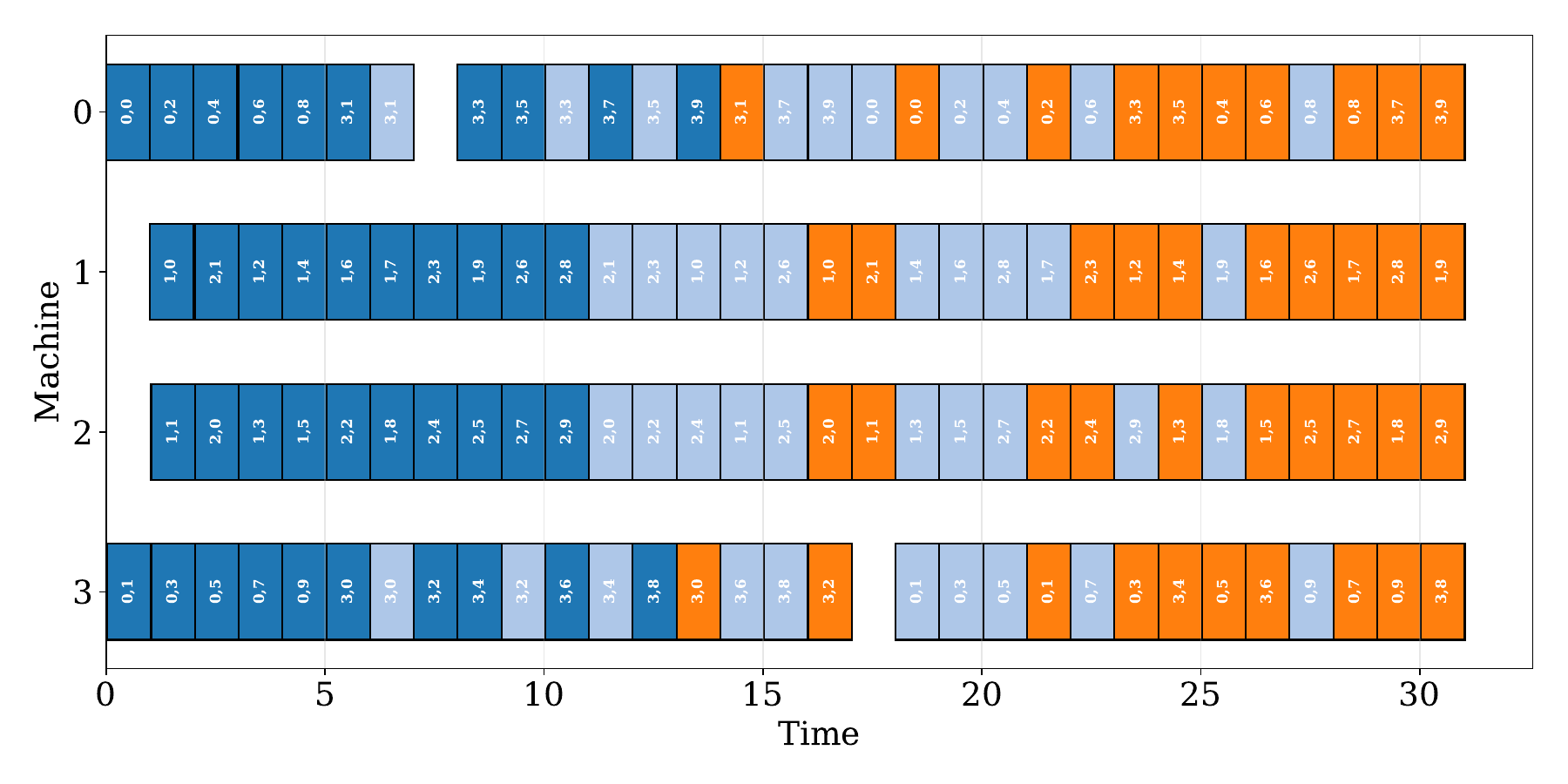} \\
  \small (b) PP-rank = 4 \\[0.5em]
  \includegraphics[width=0.98\linewidth]{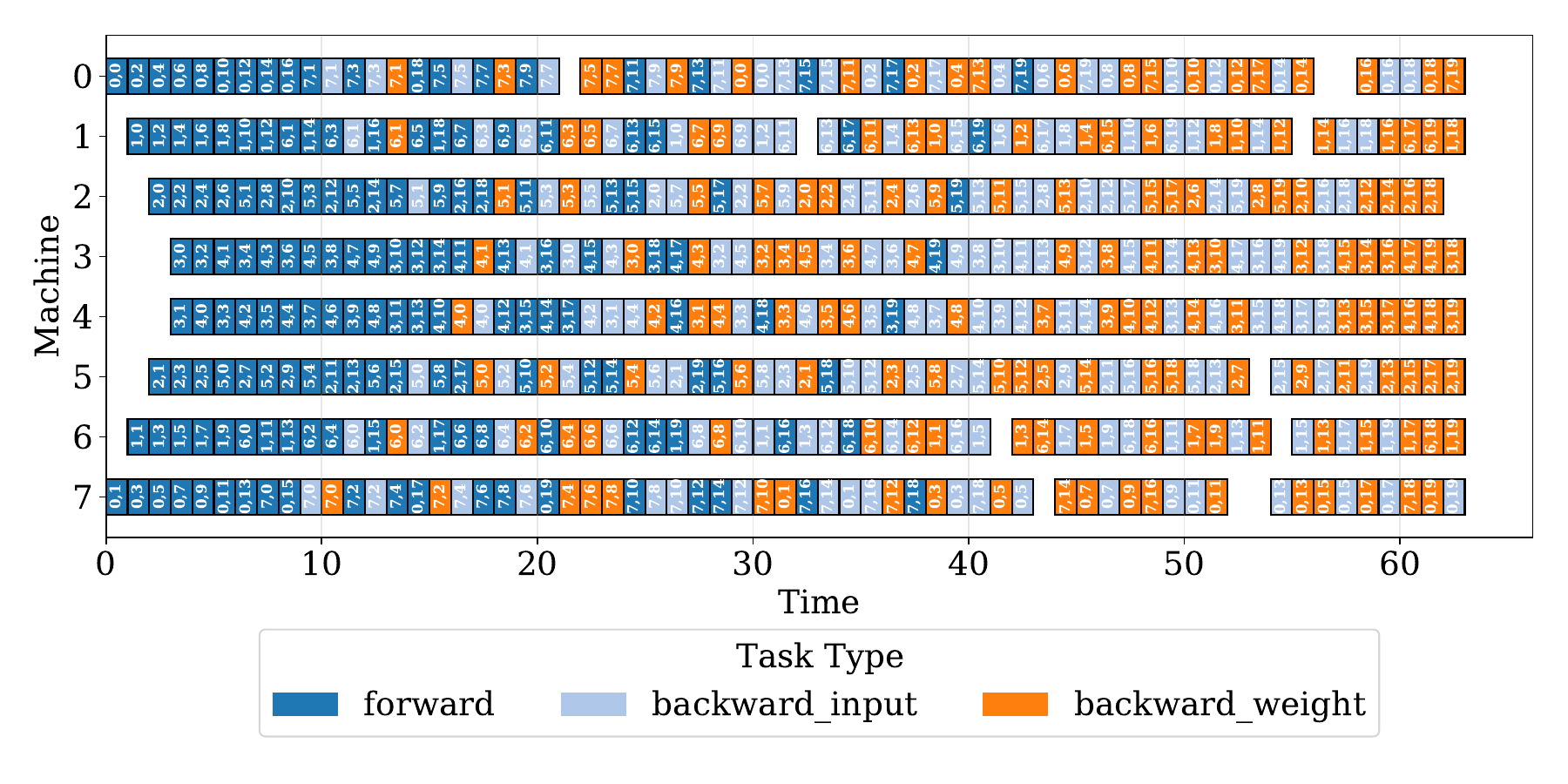} \\
  \small (c) PP-rank = 8
\end{tabular}
\caption{Gantt charts of enhanced parallelization strategies with relaxed activation memory constraints. The optimization framework discovers more regular scheduling patterns, but did not further improve bubble rate.}
\label{fig:dualpipe-relaxed-memory}
\end{figure}

While our framework can reproduce Dualpipe's strategy under identical constraints, we also investigated whether relaxing certain constraints could yield improved efficiency. We hypothesized that Dualpipe's strict memory bound might be unnecessarily conservative for certain deployment scenarios. Knowing that the model parameter requires dominant memory comparing to activation in most situations (except for extremely long sequence generation), we modified our optimization model to maintain the parameter memory allocation ($2\times$ parameters) but relaxed the activation memory constraint by doubling the allowed activation memory from $(PP + 1)$ to $2 ( PP + 1)$, representing a modest increase that remains practical for modern accelerator hardware. Figure~\ref{fig:dualpipe-relaxed-memory} shows the resulting Gantt charts with these relaxed memory constraints across the same three pipeline configurations. Surprisingly, the new schedule form a tighter and regular pattern, but achieving the same bubble size of $(PP/2-1)$ and did not improve in efficiency with the relaxed memory bound. Such a result indicates that the potential optimality of the bi-directional interweaving pattern in balancing throughput and memory efficiency.

\subsection{Random Graph}
\begin{figure}[h]
\centering
  \includegraphics[width=0.8\linewidth]{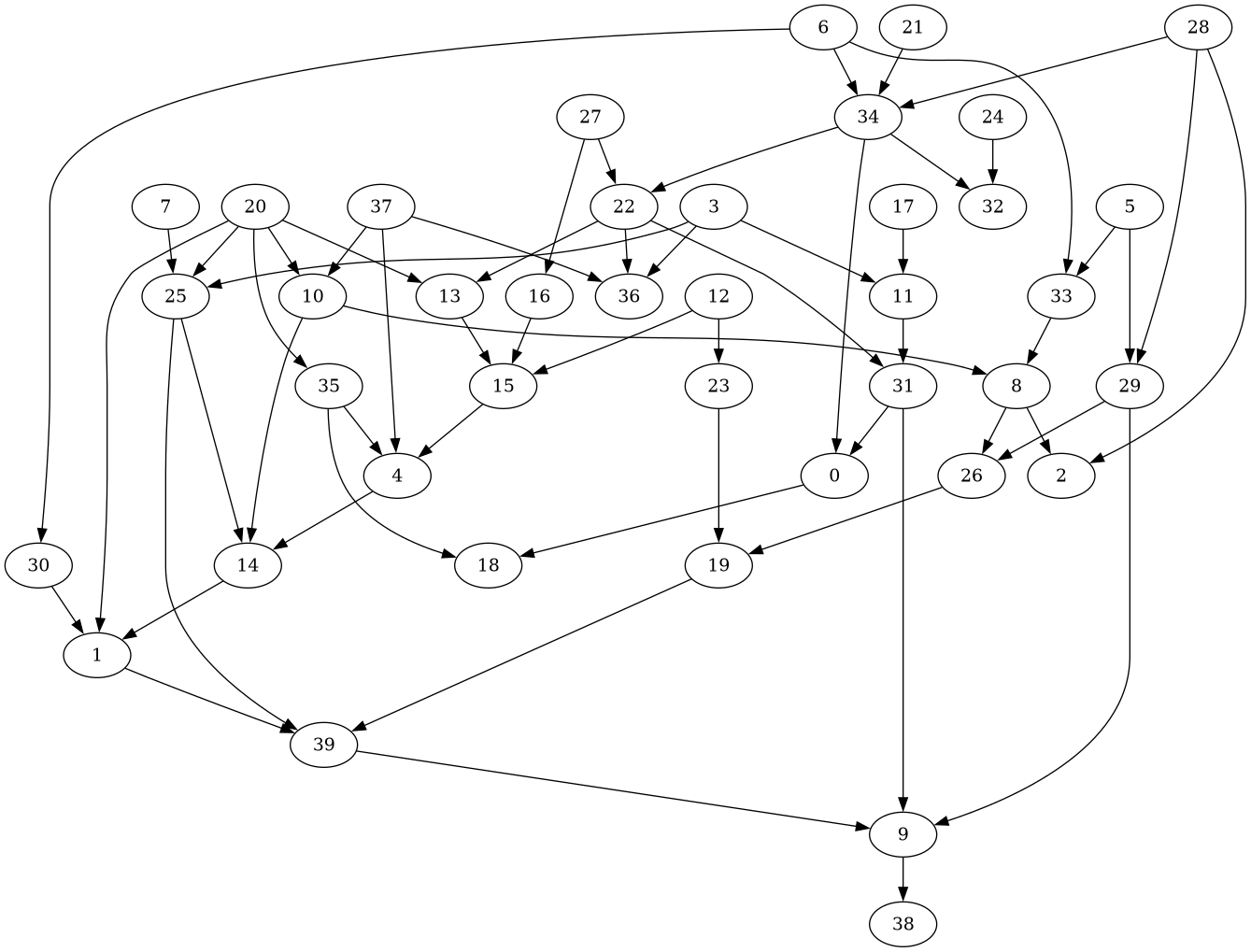}\\
  \caption{Visualization of the randomly generated computation graph coarsen from 200 nodes down to 40 nodes.}
  \label{random_graph_coarsen}
\end{figure}

\begin{figure}[h]
\centering
  \includegraphics[width=0.99\linewidth]{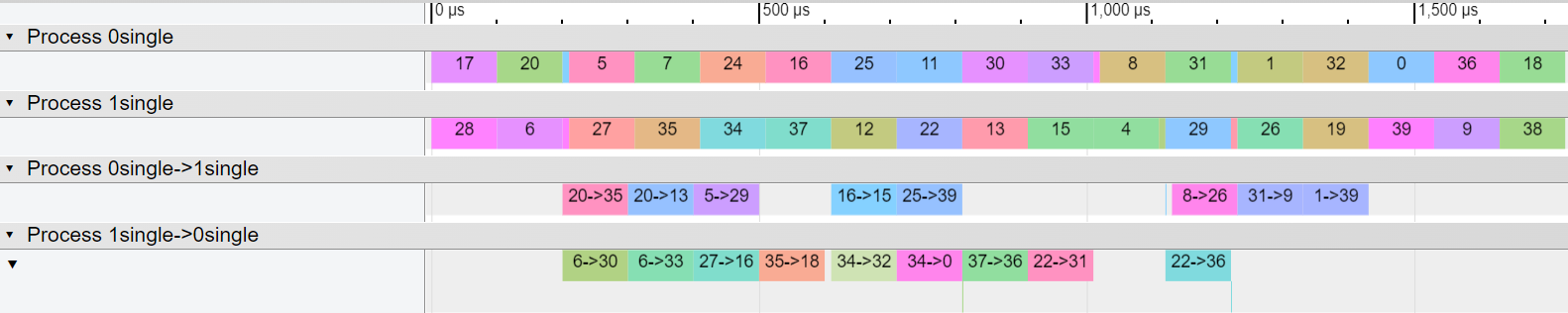}\\
  \caption{Gantt chart of an optimized network inference schedule on 2 chips}
  \label{gantt_random_graph}
\end{figure}

We now demonstrate the capability of the proposed framework to automatically parallelize a neural network with arbitrary structure. We generate a random computational graph with 200 nodes, where the maximum in- and out-degrees of nodes equal 3, representing of various types of neural network operations with randomized compute and memory requirements. We then optimize scheduling of the random computational graph on a 2-machine cluster without memory constraints.
Figure~\ref{random_graph_coarsen} shows the coarsen graph of 40 nodes yielded by Alg~\ref{algorithm1}. Figure~\ref{gantt_random_graph} presents the Gantt chart visualization utilizing the interactive trace viewing tool available at chrome://tracing. The horizontal axis represents time, while the vertical axis shows the allocation of operations to different processing units. The top two rows show the two chips for computation, and the bottom two rows show the two communication channels.
The automatic partitioning strategy generated by our optimizer perfectly balances computational load across the available hardware while respecting all dependency constraints inherent in the network structure. As evident from the visualization, our approach minimizes idle time between operations and achieves efficient pipeline parallelism, demonstrating the framework's ability to extract parallelization opportunities even in arbitrary graph structures without predefined patterns. Notably, the optimizer successfully determined the optimal sequence of dependent operations such that the communication time is fully hidden under computation, in so doing minimizing bubbles on the computation machines and minimizing total execution time. The resulting schedule achieves the theoretical maximum speedup of $2\times$ for this two-chip configuration.

\section{Conclusion and Future Work}
\label{conclusion}

This work introduces a novel mathematical programming framework that formalizes the neural network parallelization problem, providing a principled approach to automatically generate efficient distributed computation strategies. Our primary contribution lies in developing a comprehensive mathematical formulation that captures the necessary and sufficient conditions for valid parallelization plans across diverse neural network architectures and hardware configurations.

The mathematical program we have developed represents a significant advancement over existing approaches in several key dimensions. First, it offers a general formulation that accommodates arbitrary computational graphs, moving beyond the limitations of techniques designed for specific network architectures. Second, it provides a complete model that simultaneously addresses multiple aspects of parallelization—including operation placement, scheduling, memory management, and communication—within a unified optimization framework. Third, it establishes an expandable foundation that can be readily extended to incorporate additional constraints, objectives, or parallelization dimensions as required by specific deployment scenarios.

We have demonstrated the versatility of our framework through several variants that address practical deployment considerations. The memory-bounded formulation enables efficient parallelization while respecting hardware memory limitations, a critical constraint for large-scale models. The dynamic loading/unloading extension optimizes activation memory usage through strategic recomputation, allowing for more efficient execution of memory-intensive networks. Each variant maintains the mathematical rigor of the core formulation while adapting to specific operational requirements.

Our empirical evaluation confirms the effectiveness of the proposed approach across different scenarios. When applied to randomly generated computational graphs, our framework successfully identifies efficient parallelization strategies that balance computational load while minimizing communication overhead. This result highlights the advantage of our mathematical programming approach over heuristic methods, as it can discover non-obvious parallelization strategies that might be difficult to identify manually, particularly for complex and irregular network architectures where intuition-based approaches often falter.

More remarkably, our comparison with Dualpipe—an expert-designed parallelization strategy implemented in DeepSeek V3—demonstrates both the validation and innovation capabilities of our approach. The framework successfully reproduces Dualpipe's sophisticated scheduling patterns when given identical constraints, validating its ability to automatically derive strategies that previously required extensive expert engineering effort. Furthermore, by systematically exploring relaxations to memory constraints, our optimizer discovers enhanced parallelization strategies that substantially reduce pipeline bubbles and improve hardware utilization.

The enhanced parallelization strategies discovered by our framework demonstrate the value of automated optimization approaches. While expert-designed strategies like Dualpipe incorporate important domain knowledge, they often reflect specific design constraints that may not be optimal across all deployment scenarios. Our mathematical programming framework enables systematic exploration of this design space, automatically discovering improved parallelization strategies tailored to specific hardware configurations and application requirements.

This work bridges the gap between theoretical optimization and practical deployment of distributed neural networks. By formalizing parallelization as a mathematical program, we enable both researchers and practitioners to reason systematically about trade-offs in the design space, identify efficiency opportunities that may be missed by heuristic approaches, and automatically generate optimized execution plans for diverse neural network architectures.

Looking forward, the expandable nature of our framework provides a foundation for future research directions. Potential extensions include incorporating heterogeneous hardware considerations, dynamic load balancing for unpredictable workloads, and automated exploration of hybrid parallelization strategies that combine different dimensions of parallelism. As neural networks continue to grow in scale and complexity, the principled optimization approach presented in this work offers a promising path toward efficient and automated deployment across increasingly diverse computational environments.


%





\ifCLASSOPTIONcaptionsoff
  \newpage
\fi



%



\bibliographystyle{IEEEtran}
\bibliography{IEEEabrv,myBib}

\end{document}